\documentclass[utf8]{FrontiersinHarvard} 

\usepackage{url,hyperref,lineno,microtype,subcaption}
\usepackage[onehalfspacing]{setspace}

\usepackage{float}
\usepackage{booktabs}

\def\keyFont{\fontsize{8}{11}\helveticabold }
\def\firstAuthorLast{Inigo {et~al.}} 
\def\Authors{Blanca Inigo \,$^{1,*}$, Benjamin D. Killeen\,$^{1}$, Rebecca Choi\,$^{2}$, Michelle Song\,$^{1}$, Ali Uneri\,$^{2}$, Majid Khan\,$^{2}$, Christopher Bailey\,$^{2}$, Axel Krieger\,$^{1}$ and Mathias Unberath\,$^{1}$}
\def\Address{$^{1}$Johns Hopkins Whiting School of Engineering, Baltimore, MD, USA \\
$^{2}$Johns Hopkins School of Medicine, Baltimore, MD, USA  }
\def\corrAuthor{Blanca Inigo}

\def\corrEmail{binigo1@jhu.edu}

\begin{document}
\onecolumn
\firstpage{1}

\title[CT-Free 3D Path Planning from bi-plane X-rays]{3D Path Planning for Robot-assisted Vertebroplasty from Arbitrary Bi-plane X-ray via Differentiable Rendering} 

\author[\firstAuthorLast ]{\Authors} 
\address{} 
\correspondance{} 

\extraAuth{}

\maketitle

\begin{abstract}

Robotic systems are transforming image-guided interventions by enhancing accuracy and minimizing radiation exposure. A significant challenge in robotic assistance lies in surgical path planning, which often relies on the registration of intraoperative 2D images with preoperative 3D CT scans. This requirement can be burdensome and costly, particularly in procedures like vertebroplasty, where preoperative CT scans are not routinely performed. To address this issue, we introduce a differentiable rendering-based framework for 3D transpedicular path planning utilizing bi-planar 2D X-rays. Our method integrates differentiable rendering with a vertebral atlas generated through a Statistical Shape Model (SSM) and employs a learned similarity loss to refine the SSM shape and pose dynamically, independent of fixed imaging geometries. We evaluated our framework in two stages: first, through vertebral reconstruction from orthogonal X-rays for benchmarking, and second, via clinician-in-the-loop path planning using arbitrary-view X-rays. Our results indicate that our method outperformed a normalized cross-correlation baseline in reconstruction metrics (DICE: 0.75 vs. 0.65) and achieved comparable performance to the state-of-the-art model ReVerteR (DICE: 0.77), while maintaining generalization to arbitrary views. Success rates for bipedicular planning reached 82\% with synthetic data and 75\% with cadaver data, exceeding the 66\% and 31\% rates of a 2D-to-3D baseline, respectively. In conclusion, our framework facilitates versatile, CT-free 3D path planning for robot-assisted vertebroplasty, effectively accommodating real-world imaging diversity without the need for preoperative CT scans.

\tiny
 \keyFont{ \section{Keywords:} Artificial Intelligence, Deep Learning, Computer-Assisted Interventions, Intraoperative Planning, Fluoroscopy, Spine Surgery} 
\end{abstract}

\section{Introduction}

\label{sec:introduction}
\begin{figure*}[!htbp]
    \centering
    \includegraphics[width=\textwidth]{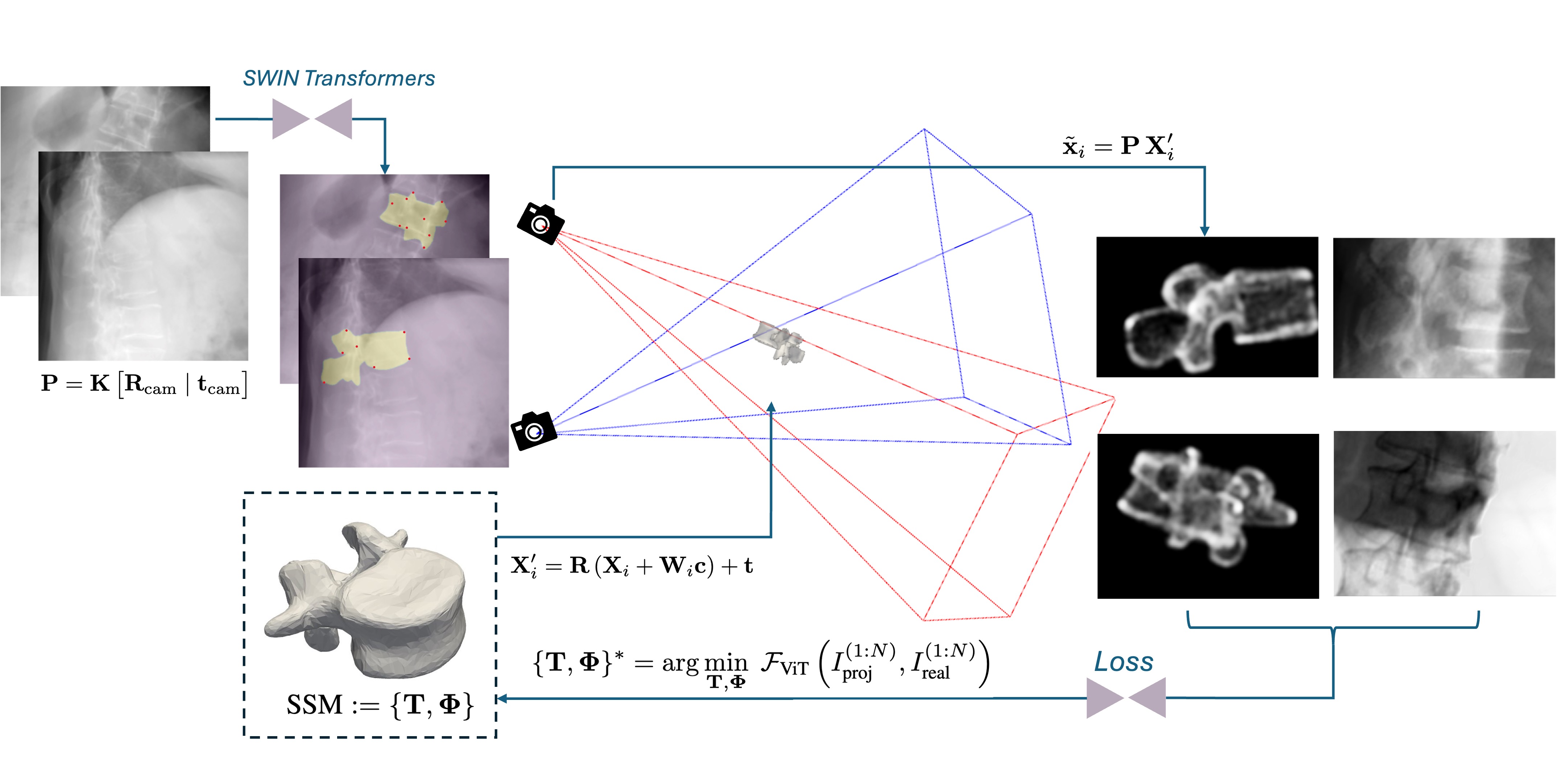} 
    \caption{Proposed pipeline overview: Two SWIN Transformers analyze X-ray images to detect and segment vertebrae, identifying landmarks to initialize the SSM's pose. Gaussian splatting is used to render SSM projections in optimization steps. A multi-view Vision Transformer (ViT) calculates a similarity loss from these projections and original X-rays, refining the SSM's pose and shape for accurate 3D vertebral reconstruction with consistent point correspondences.}
    \label{fig:pipeline}
\end{figure*}

Robotic systems are transforming the Operating Room (OR) by improving surgical outcomes, increasing safety, and reducing radiation exposure through remote actuation and fewer fluoroscopy acquisitions. A key component of robotic automation is accurate surgical path planning, which typically requires detailed 3D anatomical information. For complex spinal procedures, this information is obtained through preoperative CT scans that can be registered with intraoperative imaging to guide the robot. However, for less severe conditions such as vertebral compression fractures, pre-operative CTs are not available as they are not needed to diagnose the conditions. For these procedures, acquiring a preoperative CT scan is impractical due to increased time, cost, and radiation exposure. As a result, high-volume yet relatively straightforward procedures, like vertebroplasty, have seen limited benefit from robotic assistance. 
Developing methods that enable accurate surgical planning using just a few 2D X-ray images could make robotic assistance viable for these cases.

Over the past decade, deep learning models have shown potential for reconstructing 3D anatomy from 2D images, potentially eliminating the need for CT scans. However, these models often operate under strict assumptions, such as the availability of standardized imaging angles or minimal relative tilt between views~\cite{Kyung2023Reconst,CHEN2023106615,saravi2023synthetic,chen2024automatic}. These constraints limit their applicability in real-world clinical settings, where imaging conditions vary widely and the most informative views are often patient-specific~\cite{killeen2023autonomous}. Additionally, many of these models act as “black boxes,” lacking interpretability and explicit anatomical constraints that are particularly concerning in high-stakes applications such as spine surgery.

Vertebroplasty —a minimally invasive procedure where a needle is carefully inserted through the pedicle into the vertebral body to inject bone cement— could benefit from robotic assistance to improve safety and efficacy. However, existing systems are often bulky, costly, and disruptive to clinical workflows, leading to limited adoption, particularly in low-volume settings~\cite{DSouza2019}.

To overcome these limitations, we present \textit{Spine-DART} (Differentiable Atlas-based Reconstruction and Trajectory planning for the Spine), a needle trajectory planning strategy that uses intraoperative 2D X-ray images to generate accurate 3D guidance without requiring a preoperative CT scan. Our optimization-based framework performs end-to-end 3D transpedicular path planning from two arbitrary X-rays. By combining differentiable rendering with a vertebral Statistical Shape Model (SSM) and a learned similarity loss, the method jointly optimizes the SSM’s pose and shape without assuming fixed imaging geometry. This allows for anatomically plausible shape constraints and consistent point correspondences, enabling effective alignment of 3D anatomy and planned paths with the input X-rays.

Unlike existing deep learning methods that rely on fixed imaging conditions, Spine-DART adapts to the variability of real-world intraoperative fluoroscopy, enabling generalizable 3D planning from arbitrary views. While atlas-based pipelines offer valuable anatomical priors, they often depend on hand-crafted similarity metrics and non-differentiable optimization procedures that are sensitive to initialization and prone to failure without manual intervention. These limitations hinder their robustness, scalability, and integration into automated systems.

Spine-DART overcomes these challenges by integrating statistical shape models with deep learning. By combining anatomical priors with a learned similarity loss and differentiable rendering, our framework supports gradient-based optimization of shape and pose, yielding anatomically consistent reconstructions without requiring preoperative CT or constrained X-ray geometries. This hybrid approach balances interpretability with flexibility, enabling accurate and reliable CT-free 3D path planning in unconstrained intraoperative settings.

By eliminating the need for CT and integrating seamlessly into current clinical workflows, Spine-DART enhances surgical planning precision and supports the broader adoption of compact, cost-effective robotic platforms for procedures like vertebroplasty~\cite{9635356,11025971}.

\section{Related work}

3D surgical path planning, particularly for transpedicular access, has traditionally relied on preoperative CT-based methods, which provide detailed anatomical models for direct 3D planning and integration with navigation systems~\cite{otomo2022ctnavigation,9722904,singh2020ctbiopsy,wang2024deep}. In intraoperative scenarios where CT is unavailable or impractical, alternative strategies have emerged. These can be broadly categorized as: (1) direct 2D path planning followed by triangulation or interpolation to recover a 3D trajectory~\cite{Killeen2023Pelphix,Naik2022HybridRegistration}, and (2) 3D anatomical reconstruction from intraoperative 2D images followed by path planning~\cite{YANG2024108444,Shiode2021ForearmReconstruction,Uneri2014RegistrationViewAngles}. While both approaches eliminate the need for preoperative imaging, they present trade-offs in terms of accuracy, sensitivity to imaging geometry, workflow integration, and interpretability. Our method extends the second category by enabling 3D vertebral reconstruction and automatic path planning from any two fluoroscopic viewpoints without requiring known geometries or standardized angles.

The first category, 2D path planning with 3D interpolation, involves annotating surgical paths directly on fluoroscopic images and then reconstructing the 3D trajectory via triangulation. Killeen et al.~\cite{killeen2023autonomous} demonstrated this technique using fluoroscopy images for pelvic screw fixation. While this method is straightforward and computationally efficient, the 2D predictions are made independently for each view, without considering their correlation through the underlying 3D path. This can result in misaligned annotations that reduce the accuracy of the final surgical plan.

The second category reconstructs 3D anatomy from 2D images, allowing path planning directly in 3D space. Deep learning (DL) methods have shown promise in this area for their reconstruction accuracy and inference speed~\cite{chen2024automatic}. Approaches by Kyung et al.~\cite{Kyung2023Reconst}, Chen et al.~\cite{CHEN2023106615}, and Ying et al.~\cite{ying2019x2ct} require paired orthogonal X-rays (e.g., AP and lateral) to constrain the reconstruction, limiting their flexibility in unconstrained intraoperative settings. Furthermore, most methods are trained and tested on Digitally Reconstructed Radiographs (DRRs), which can lead to generalization challenges when applied to real X-ray images.

Atlas-based methods offer an alternative by deforming a Statistical Shape Model to match input images.  These ensure anatomical plausibility and enable automatic propagation of annotations~\cite{Vijayan2019AutoPlan, Han2019, Lamecker2006AtlasBased, Ehlke2020}. However, traditional atlas-based registration pipelines typically rely on conventional optimization techniques, which are sensitive to initialization and often exhibit a limited capture range. As a result, they generally require semi-manual interaction to ensure convergence to anatomically valid solutions, limiting their scalability in time-sensitive surgical settings~\cite{Vijayan2019PediclePlanning, Vijayan2019AutoPlan, Yousefi2011, 7080645}. Moreover, their non-differentiable components pose challenges for integration into modern learning-based frameworks. 

Spine-DART addresses these challenges by combining atlas priors with deep learning, preserving the anatomical plausibility and interpretability of SSMs while overcoming the limitations of conventional optimization. By incorporating differentiable rendering and a learned similarity loss, Spine-DART enables joint optimization of shape and pose through gradient-based methods. This framework supports robust 3D path planning from arbitrary X-ray viewpoints without relying on patient CT or constrained imaging geometry, while also maintaining interpretability through intermediate reconstruction and planning steps.

\section{Methods}

Spine-DART enables 3D vertebral reconstruction and automatic transpedicular path planning directly from biplanar 2D X-ray images (Figure \ref{fig:pipeline}). It consists of four main components:  
\begin{enumerate}
    \item \textbf{Vertebral SSMs} that act as shape priors, with their pose and shape parameters optimized during registration.
    \item \textbf{DL networks} that extract semantic features from X-rays to initialize the SSM pose. 
    \item \textbf{A differentiable rendering pipeline} based on Gaussian splatting that generates vertebral projections while supporting gradient-based optimization.
    \item \textbf{A multi-view learned similarity loss} between the rendered projections and the input X-rays.
\end{enumerate}

Since the vertebral SSM is pre-annotated with the transpedicular trajectory, the final surgical path plan is inherently transferred to the registered SSM during optimization.

\subsection{Vertebral Statistical Shape Model}

Statistical Shape Models (SSMs) of anatomical structures have long supported surgical planning and automated image annotation~\cite{Killeen2023Pelphix, Ambellan2019SSM}. In our approach, SSMs serve two key purposes: they reduce optimization complexity by constraining the solution space to anatomically plausible vertebral shapes, and they enable direct propagation of anatomical annotations—such as transpedicular trajectories—through consistent point correspondences. Given the substantial anatomical variability along the spine, we follow common practice and construct region-specific SSMs for different vertebral groups to improve both specificity and anatomical realism~\cite{Panjabi1980, Bredbenner2014CervicalSSM, Wai2023ThoracicSSM}

Accordingly, we developed four distinct SSMs: (1) T1–T4, (2) T5–T8, (3) T9–T12, and (4) L1–L5. These models were created using vertebral segmentations from the CTSpine1K dataset~\cite{deng2023ctspine1k}, excluding subjects from the VerSe dataset~\cite{verse19}. To generate the SSMs, we first performed rigid registration to establish point correspondences across shapes, followed by multiple iterations of deformable registration and Principal Component Analysis (PCA). This process ensured that the resulting models captured both typical anatomical features and natural inter-subject variability while excluding implausible deformations. 

Each resulting SSM is parameterized by 15 principal modes of variation, which account for approximately 90\% of the population-level shape variability. These shape parameters are jointly optimized with 6 pose degrees of freedom (3 for rotation and 3 for translation) during the reconstruction process.

\subsection{Automatic X-ray Annotation for Semantic Contextualization}

\begin{figure}[H]
    \centering
    \includegraphics[width=0.5\linewidth]{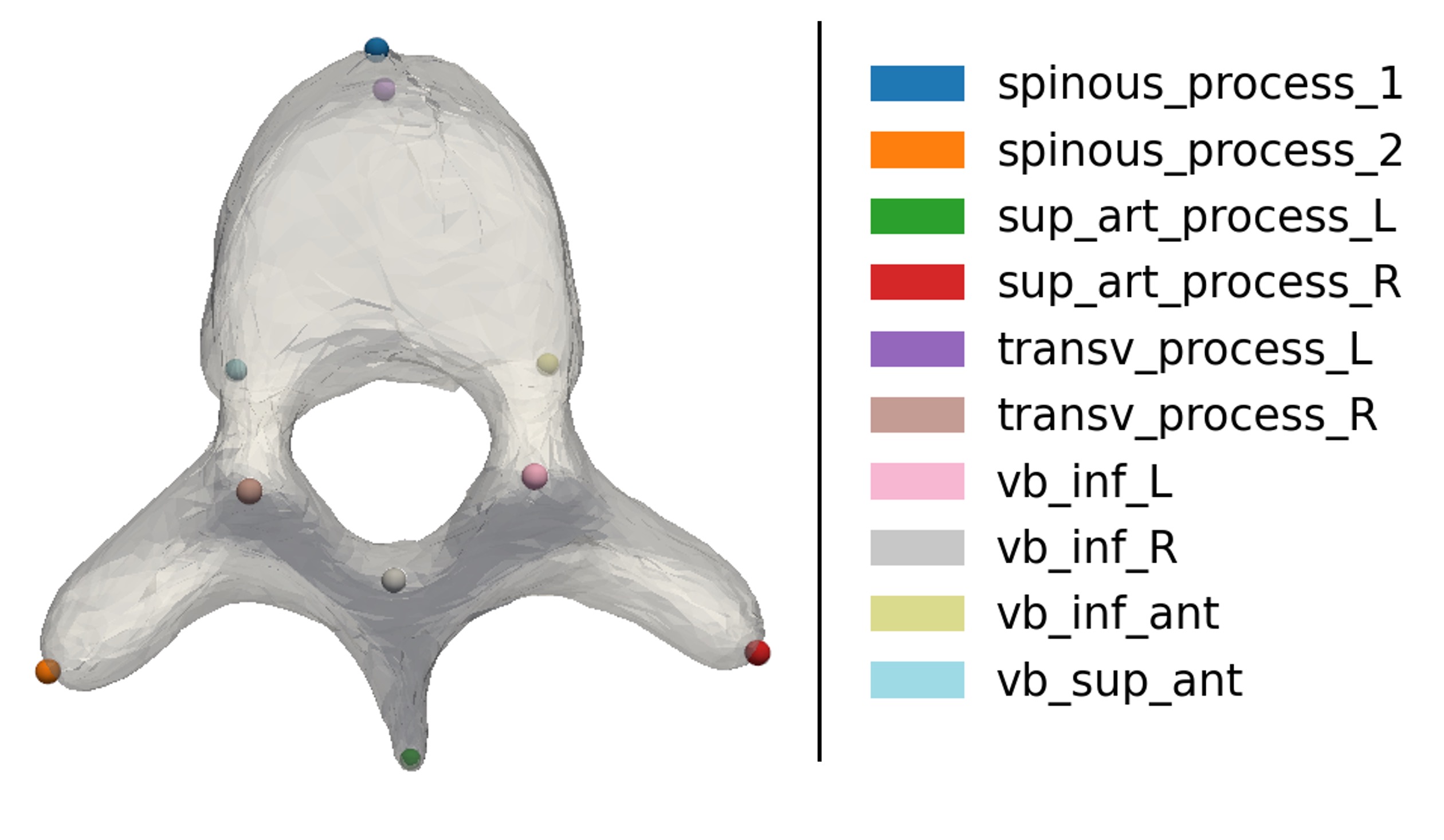}
    \caption{3D vertebral landmarks used for initialization of the SSM 3D pose.}
    \label{fig:vert_landmarks}
\end{figure}

Like most optimization-based methods, Spine-DART requires a reasonably accurate initialization of parameters. To achieve this, we trained two DL models with the same SWIN Transformer backbone~\cite{liu2021Swin} to perform vertebral semantic segmentation and landmark detection on X-ray images. Both models were trained using Digitally Reconstructed Radiographs (DRRs) generated under controlled conditions, with sufficient variability in imaging parameters and domain randomization to bridge the gap between synthetic and real X-rays~\cite{gao2023synthetic}. Additionally, we leveraged our SSMs to propagate 10 anatomical landmarks within the vertebral body and processes (Figure \ref{fig:vert_landmarks}). To ensure high-quality supervision, we included only those vertebrae where the SSM achieved a registration DICE score above 85\%, indicating a reliable fit.

Using DeepDRR~\cite{Unberath2018DeepDRR, DeepDRR2019, Gao2023SyntheX}, we generated a training dataset of over 150,000 DRRs, each paired with its corresponding 2D vertebral segmentations and keypoints. 

\subsection{Differentiable Rendering for 3D Vertebra Reconstruction}

Differentiable rendering simulates the generation of 2D images from 3D models while enabling gradient computation and backpropagation. This allows for the optimization of 3D parameters by comparing rendered projections to input X-rays. Our approach leverages this framework to iteratively optimize vertebral \textbf{shape} and \textbf{pose} parameters. Each component in the rendering pipeline is differentiable with respect to the 21 degrees of freedom (DoF), which are split into:

\begin{itemize}
    \item \textbf{Rigid transformation} \( \mathbf{T} = \{\mathbf{R}, \mathbf{t}\} \): 6 DoF, where \( \mathbf{R} \in \mathrm{SO}(3) \) is a rotation matrix and \( \mathbf{t} \in \mathbb{R}^3 \) is a translation vector.
    \item \textbf{Shape deformation} \( \boldsymbol{\Phi} \): 15 DoF, parameterized by shape coefficients \( \mathbf{c} \in \mathbb{R}^{15} \) associated with the principal modes of variation of the Statistical Shape Model (SSM).
\end{itemize}

We define the SSM parameterization as:
\[
\text{SSM} := \{\mathbf{T}, \boldsymbol{\Phi}\}
\]

Each new shape is defined as a linear combination of the shape basis followed by a rigid transformation. Specifically, for each canonical vertex \( \mathbf{X}_i \in \mathbb{R}^3 \), the transformed vertex is given by:
\[
\mathbf{X}_i' = \mathbf{R} \left( \mathbf{X}_i + \mathbf{W}_i \mathbf{c} \right) + \mathbf{t}
\]
where \( \mathbf{W}_i \in \mathbb{R}^{3 \times 15} \) is the row-block of the shape basis matrix \( \mathbf{W} \in \mathbb{R}^{3N \times 15} \) corresponding to vertex \( i \), and \( \mathbf{c} \in \mathbb{R}^{15} \) are the shape coefficients.

This formulation allows for continuous and differentiable shape deformation and pose transformation.

To render the transformed SSM in 2D, we use a CUDA-accelerated Gaussian Splatting renderer~\cite{ye2024gsplatopensourcelibrarygaussian}, which models each vertex as an anisotropic 3D Gaussian. Given a pinhole camera projection matrix:
\[
\mathbf{P} = \mathbf{K} \begin{bmatrix} \mathbf{R}_\mathrm{cam} \mid \mathbf{t}_\mathrm{cam} \end{bmatrix},
\]
each transformed vertex \( \mathbf{X}_i' \in \mathbb{R}^4 \) (in homogeneous coordinates) is projected into the image plane as:
\[
\tilde{\mathbf{x}}_i = \mathbf{P} \, \mathbf{X}_i',
\quad \text{and} \quad
\mathbf{x}_i = \left( \frac{\tilde{x}_i}{\tilde{z}_i}, \frac{\tilde{y}_i}{\tilde{z}_i} \right)
\]

Gaussian Splatting renders the 2D projections \( \mathbf{x}_i \) as elliptical blobs, accounting for local surface orientation and perspective. This enables end-to-end differentiability and gradient flow through both shape and pose parameters.

To address discrepancies between real and generated projections due to differences in image formation, we introduce a learned similarity metric, \( \mathcal{L}_{\text{sim}} \), based on a multi-view Vision Transformer (ViT)~\cite{10483891}, which is more robust than traditional metrics such as mutual information (MI) or normalized cross-correlation (NCC). The similarity loss is defined as:
\[
\mathcal{L}_{\text{sim}} = \mathcal{F}_{\text{ViT}} \left( I_{\text{proj}}^{(1:N)}, I_{\text{real}}^{(1:N)} \right)
\]
where \( I_{\text{proj}}^{(1:N)} \) and \( I_{\text{real}}^{(1:N)} \) are the rendered and real X-ray images from \( N \) viewpoints (with \( N = 2 \) in our setup), and \( \mathcal{F}_{\text{ViT}} \) encodes spatial and geometric consistency between views.

The goal is to find the optimal shape and pose parameters that minimize the similarity loss:
\[
\{\mathbf{T}, \boldsymbol{\Phi} \}^* = \arg \min_{\mathbf{T}, \boldsymbol{\Phi}} \; \mathcal{F}_{\text{ViT}} \left( I_{\text{proj}}^{(1:N)}, I_{\text{real}}^{(1:N)} \right)
\]

This differentiable formulation enables end-to-end gradient-based optimization. The similarity loss \( \mathcal{L}_{\text{sim}} \) is backpropagated through the ViT, the Gaussian splatting renderer, and the SSM parameters, allowing iterative refinement of both vertebral shape \( \boldsymbol{\Phi} \) and pose \( \mathbf{T} \) until the rendered projections align with the real X-rays.

\subsection{Automatic Transpedicular Path Planning}

Finally, having previously annotated the transpedicular entry and exit points in our vertebral SSMs, the path is inherently transferred to the reconstructed vertebra.

\section{Experimental Results}

\subsection{Dataset}

The trainable components of Spine-DART —including the Swin Transformers for semantic segmentation and landmark detection, and the Vision Transformer (ViT) for the learned similarity metric— were trained on DRRs generated from two sources. First, we used 280 CT scans from the colon subset of the CTSpine1K dataset~\cite{Spine1kDeng2024}, selected for their relatively small slice thickness ($0.88 \pm 0.16$ mm) compared to the other CTSpine1K cohorts, which is important for generating high-fidelity DRRs that closely resemble real X-ray images. Second, we incorporated 97 high-resolution torso CT scans from the New Mexico Decedent Image Database (NMDID)~\cite{NMDID} to further enrich anatomical diversity and image resolution.

For evaluation, we used the VerSe-small dataset described in~\cite{chen2024automatic} as our benchmark and compared our vertebra reconstruction results with the ReVerteR model. The final test set consisted of 140 CT scans comprising a total of 1407 vertebrae.

We generated two DRR datasets~\cite{Unberath2018DeepDRR} from VerSe-small for evaluation:
\begin{enumerate}
    \item \textit{VerSe-small\_ort}: A set containing only orthogonal views —Anteroposterior (AP) and Lateral (LAT)— to enable direct comparison with the reconstruction results reported by ReVerteR.
    \item \textit{VerSe-small\_random}: A second set consisting of uniformly distributed DRRs, synthesized around the spinal axis of each scan to evaluate our method under more diverse view conditions.
\end{enumerate}

In addition to synthetic evaluations, we tested our model on real X-ray data acquired from a cadaver study. For this experiment, we used a LoopX device to collect a CT scan and 35 fluoroscopic images from varying orientations. Ground-truth vertebra segmentations were obtained using TotalSegmentator~\cite{wasserthal2023totalsegmentator} with manual refinement to ensure accuracy. Table \ref{tab:dataset-overview} shows an overview of the data used.

\begin{table}[t]
    \centering
    \small 
    \setlength{\tabcolsep}{8pt} 
    \renewcommand{\arraystretch}{0.6} 
    \begin{tabular}{lcccc}
        \toprule
        & \multicolumn{2}{c}{{Training}} & \multicolumn{2}{c}{{Testing}} \\
        \cmidrule(lr){2-3} \cmidrule(lr){4-5}
        & \multicolumn{1}{c}{\small {colon}} & \multicolumn{1}{c}{{NMDID}} & \multicolumn{1}{c}{{VerSe-small}} & \multicolumn{1}{c}{{Specimen}} \\
        \midrule
        {CT}       & 280 & 97 & 140 & 1 \\
        {Vertebrae}& 4036 & 1530 & 1407 & 16 \\
        {DRR/X-ray}& 113,000 & 48,000 & 7000 & 35 \\
        \bottomrule
    \end{tabular}
    \caption{Overview of datasets used for training and testing our DL models.}
    \label{tab:dataset-overview}
\end{table}

All our SSMs were created from the vertebra segmentations provided in the CTSpine1k dataset~\cite{Spine1kDeng2024}.

\subsection{Automatic X-ray Annotation Models}

\begin{figure}[H]
    \centering
    \small

    \raisebox{0.15\height}{ 
    \begin{minipage}[t]{0.5\textwidth}
        \vspace{0pt}
        \centering
        \scriptsize
        \setlength{\tabcolsep}{4pt}
        \renewcommand{\arraystretch}{1.0}
        \resizebox{0.8\linewidth}{!}{%
        \begin{tabular}{lccccc}
            \toprule
            { } & Precision & Recall & F1 & Dice & IoU \\
            \midrule
            DRRs   & 1.00 & 1.00 & 1.00 & 0.89 & 0.81 \\
            X-rays & 0.97 & 1.00 & 0.99 & 0.82 & 0.73 \\
            \bottomrule
        \end{tabular}
        }
    \end{minipage}
    }
    \hspace{10pt}
    \begin{minipage}[t]{1\textwidth}
        \vspace{0pt}
        \centering
        \includegraphics[width=\linewidth]{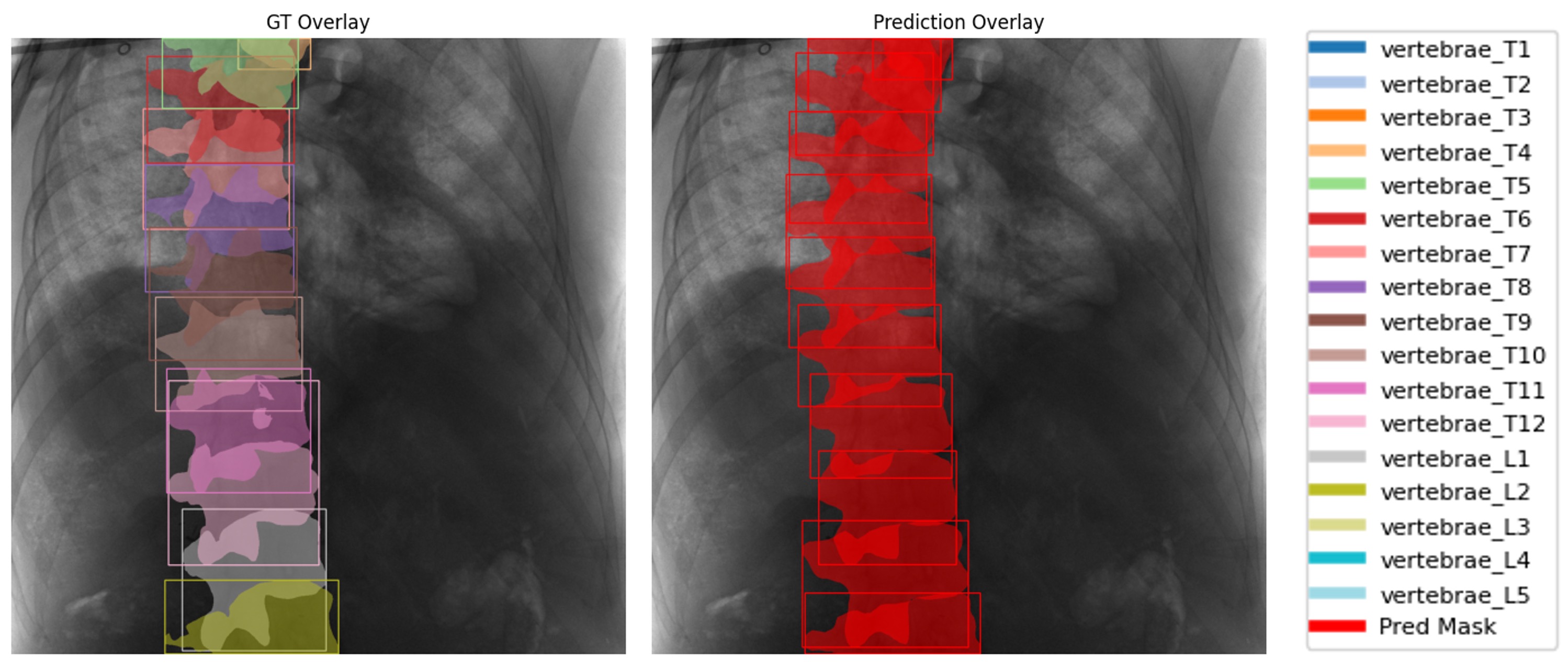}
    \end{minipage}

    \caption{Top: segmentation performance metrics for synthetic DRRs and real X-rays.  
             Right: qualitative vertebra segmentation result on a real X-ray.}
    \label{fig:metrics_qualitative}
\end{figure}

\begin{figure}
    \centering
    \includegraphics[width=0.9\linewidth]{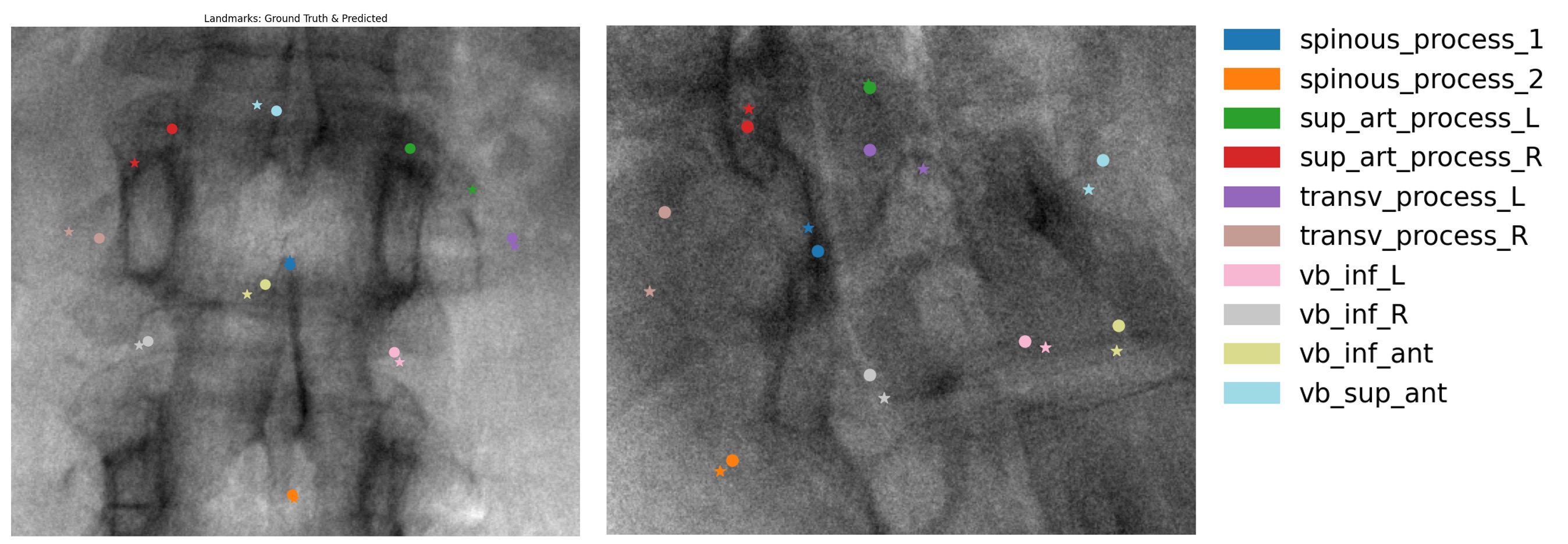}
    \caption{Qualitative results of our landmark detection model. Predictions (circles) and ground truths (stars) overlaid on real X-ray images.}
    \label{fig:qualitative_results_lmrk_det}
\end{figure}

\begin{figure}
    \centering
    \includegraphics[width=1\linewidth]{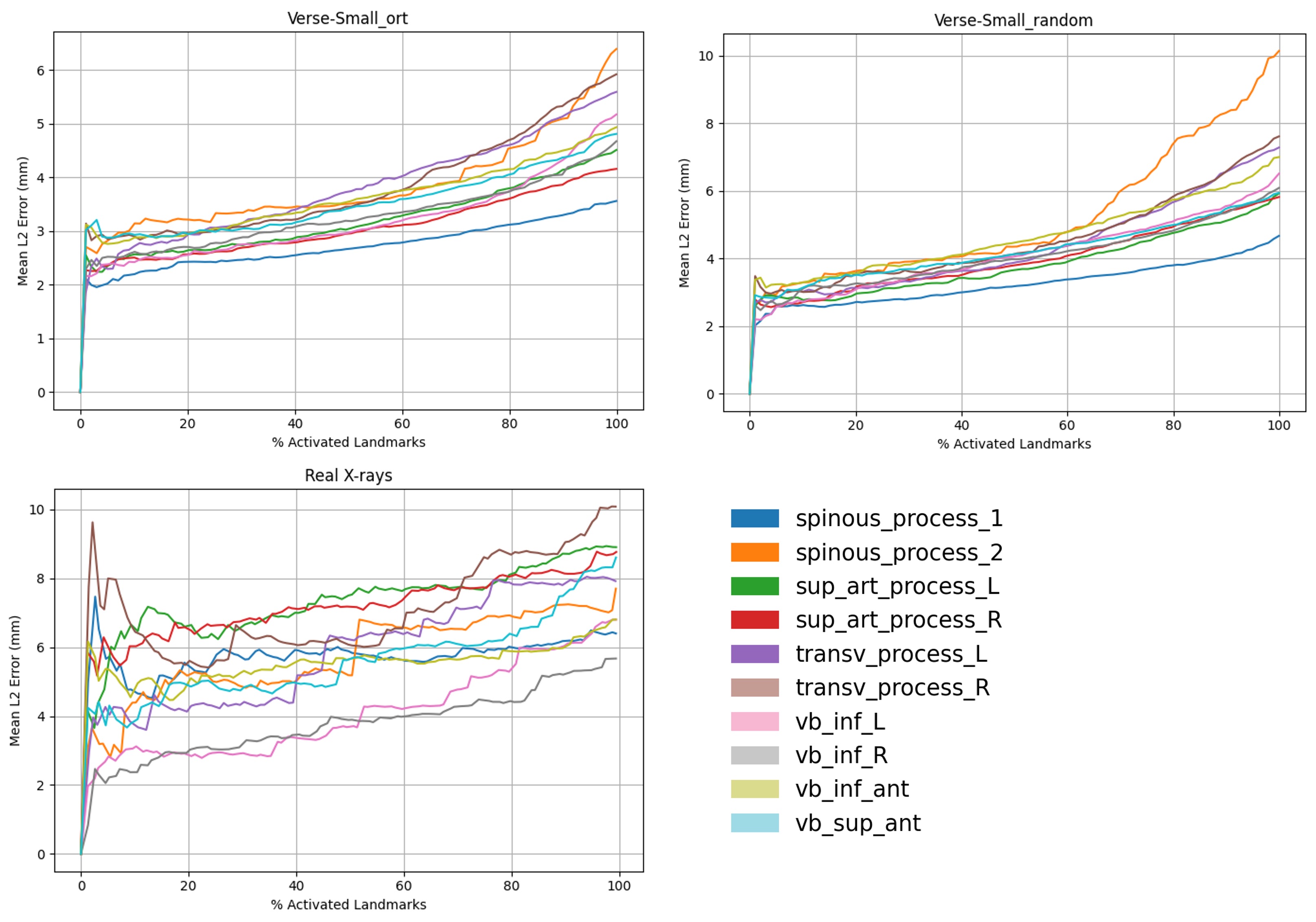}
    \caption{Landmark detection results across different domains. We report the mean Euclidean distance (mm) versus confidence percentile for orthogonal (AP/LAT) (left), random (middle), and real X-ray views (right).}
    \label{fig:landmark-metrics}
\end{figure}

We used a Mask R-CNN with a SWIN transformer backbone~\cite{mmdetection} to train a model on vertebra detection and semantic segmentation (Figure \ref{fig:metrics_qualitative}). The same backbone was retrained for landmark detection (Figure \ref{fig:qualitative_results_lmrk_det}). Both models were evaluated on VerSe-small (orthogonal and random views) and on real X-rays. Figure \ref{fig:landmark-metrics} illustrates the 2D landmark localization error (in mm) as a function of the percentage of predicted activation peaks, determined by varying the detection threshold. As more landmarks are detected (i.e., higher percentages), the average localization error increases, reflecting a trade-off between detection sensitivity and precision. 

\subsection{3D Vertebral Reconstruction}

We evaluated the accuracy of our 3D vertebra reconstruction pipeline on the VerSe-small dataset, which includes 140 spines and 1407 vertebrae. Our approach was benchmarked against ReVerteR~\cite{chen2024automatic}, a DL-based vertebra reconstruction model, using the same dataset and view protocol (AP/LAT) for fair comparison.

Each vertebra was reconstructed using two DRRs randomly selected from the synthetically generated AP and LAT views in the \textit{VerSe-small\_ort} set. Initialization was performed using vertebra detections and 2D landmarks predicted by our model, which defined the initial pose of the SSM. The reconstruction was then iteratively refined via gradient-based optimization, driven by the predicted Mean Average Surface Distance (MASD) between the rendered and real projections. This similarity loss, \( \mathcal{L}_{\text{sim}} \),  was computed using our multi-view ViT and backpropagated through the differentiable renderer to update shape and pose.

To accurately initialize the SSM pose, each pair of input views must contain at least three predicted landmarks in common to enable reliable 3D triangulation and subsequent alignment to the atlas. This condition is not always satisfied, which limits the initial reconstruction stage to 1,038 vertebrae. However, since the AP and LAT views follow fixed orientations, we can incorporate spatial priors to enable a fallback strategy. Specifically, we use a set of “backup landmarks” —including the vertebral centroid, superior, inferior, left, and right points— which allow us to initialize and reconstruct the remaining 369 vertebrae.

Table \ref{fig:reconstruction-benchmark}a presents a quantitative comparison of Spine-DART with ReVerteR~\cite{chen2024automatic} and a baseline method driven by normalized cross-correlation (NCC). While ReVerteR achieves the highest reconstruction accuracy, our optimization-based method remains competitive, outperforming the traditional NCC baseline by a large margin across all metrics. Notably, our framework offers several key advantages: it generalizes to arbitrary input views without retraining, supports integration with gradient-based priors and constraints, and maintains interpretability through its use of an SSM. This enables direct point correspondences with anatomical atlases and facilitates downstream tasks such as surgical planning or semantic labeling.

\begin{figure}[H]
    \centering
    \setlength{\tabcolsep}{8pt}
    \resizebox{0.6\textwidth}{!}{%
    \begin{tabular}{lccc}
        \toprule
        Model & DICE↑ & NSD↑ & HD-95↓ \\
        \midrule
        ReVerteR~\cite{chen2024automatic} & 0.7685 & 0.6198 & 6.3129 \\
        NCC-driven reconstruction & 0.6495 & 0.4320 & 9.2430 \\
        Spine-DART & 0.7461 & 0.6083 & 6.3431 \\
        \bottomrule
    \end{tabular}
    }
    \caption{Quantitative comparison of vertebra reconstruction accuracy across models given orthogonal-view synthetic inputs. Metrics reported are DICE, Normalized Surface Dice (NSD), and 95th percentile Hausdorff Distance (HD-95).}
    \label{fig:reconstruction-benchmark}
\end{figure}

\begin{figure}[H]
    \centering
    \includegraphics[width=0.7\textwidth]{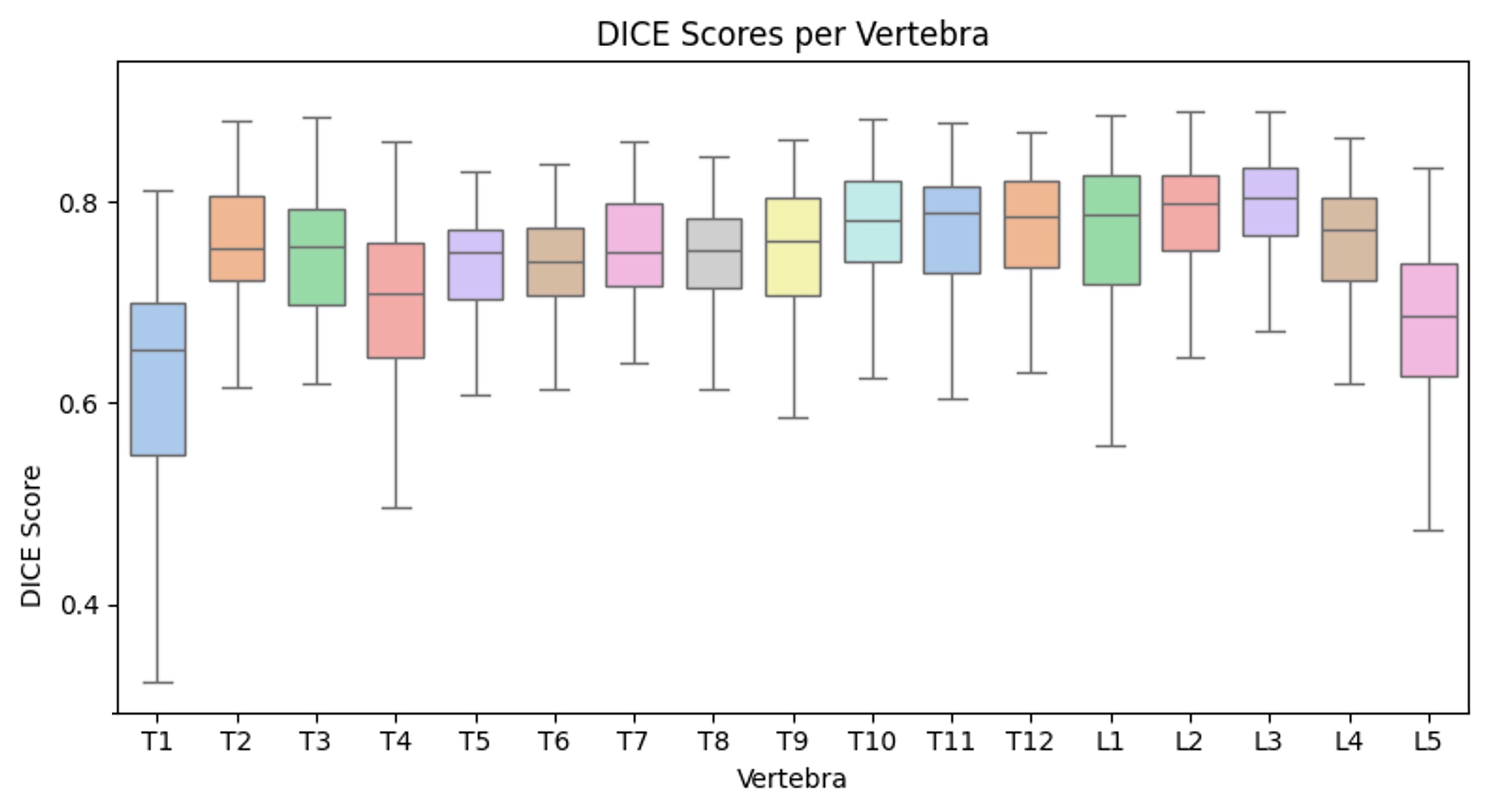}
    \caption{Per-vertebra reconstruction accuracy for Spine-DART on orthogonal-view synthetic data.}
    \label{fig:reconstruction-results}
\end{figure}

Additionally, we evaluated Spine-DART on the \textit{VerSe-small\_random} dataset to assess reconstruction performance under more diverse viewing conditions, using arbitrary pairs of randomly sampled DRRs. A total of 1,153 vertebrae were successfully reconstructed, meeting the requirement that each view pair contains at least three predicted landmarks in common. Figure \ref{fig:reconstruction-random-table} illustrates the reconstruction results on both the \textit{VerSe-small\_random} dataset and real X-rays from our cadaver specimen. Notably, 12 out of 16 vertebrae from the cadaver were reconstructed and annotated with both transpedicular trajectories, highlighting the method's applicability beyond synthetic data.

\begin{figure}[H]
    \centering
    \scriptsize
    \setlength{\tabcolsep}{10pt}
    \begin{tabular}{lccc}
        \toprule
        Dataset & DICE↑ & NSD↑ & HD-95↓ \\
        \midrule
        \textit{VerSe-small\_random} & 0.7264 & 0.5671 & 7.3337 \\
        Real X-rays & 0.7280 & 0.5231 & 7.1910 \\
        \bottomrule
    \end{tabular}
    \caption{3D vertebra reconstruction performance of our approach using arbitrary-view X-rays. Results are shown on synthetic DRRs from the \textit{VerSe-small\_random} dataset and real clinical X-rays.}
    \label{fig:reconstruction-random-table}
\end{figure}

\subsection{Automatic Transpedicular Path Planning}

While vertebra reconstruction is an important intermediate step in our pipeline —enhancing interpretability and imposing anatomical constraints— the primary goal is to automatically plan surgical paths along the vertebral pedicles. To benchmark the performance of Spine-DART, we also implement a baseline method inspired by~\cite{Killeen2023Pelphix}, which performs 2D path planning followed by geometric triangulation to estimate the 3D trajectory.

The baseline model, which we will refer to as 2D-GeoPlan, is a DL model that shares the same architecture as our vertebra segmentation network —Mask R-CNN with a SWIN backbone~\cite{mmdetection}— and was also trained on DRRs generated from the CTSpine1K and NMDID datasets. Ground truth annotations were propagated using our SSM, limited to vertebrae that were successfully registered. Trajectory segmentations were defined in 3D as cylinders with a 5 mm diameter, following~\cite{baroud2005new}. To avoid using annotations with cortical breaches, we excluded trajectories whose minimum distance to the pedicle walls was less than 2.5mm.

We applied 2D-GeoPlan to both the \textit{VerSe-small\_random} dataset and the real X-rays. As before, two views were randomly selected per sample, and the predicted paths were triangulated to generate the 3D cylindrical reconstructions. Using this approach, both transpedicular paths were successfully reconstructed for 928 out of 1,407 vertebrae in the \textit{VerSe-small\_random} dataset. For the real X-rays, full reconstruction of both paths was achieved in 5 out of 16 vertebrae. 

\begin{figure}
    \centering
    \includegraphics[width=1\textwidth]{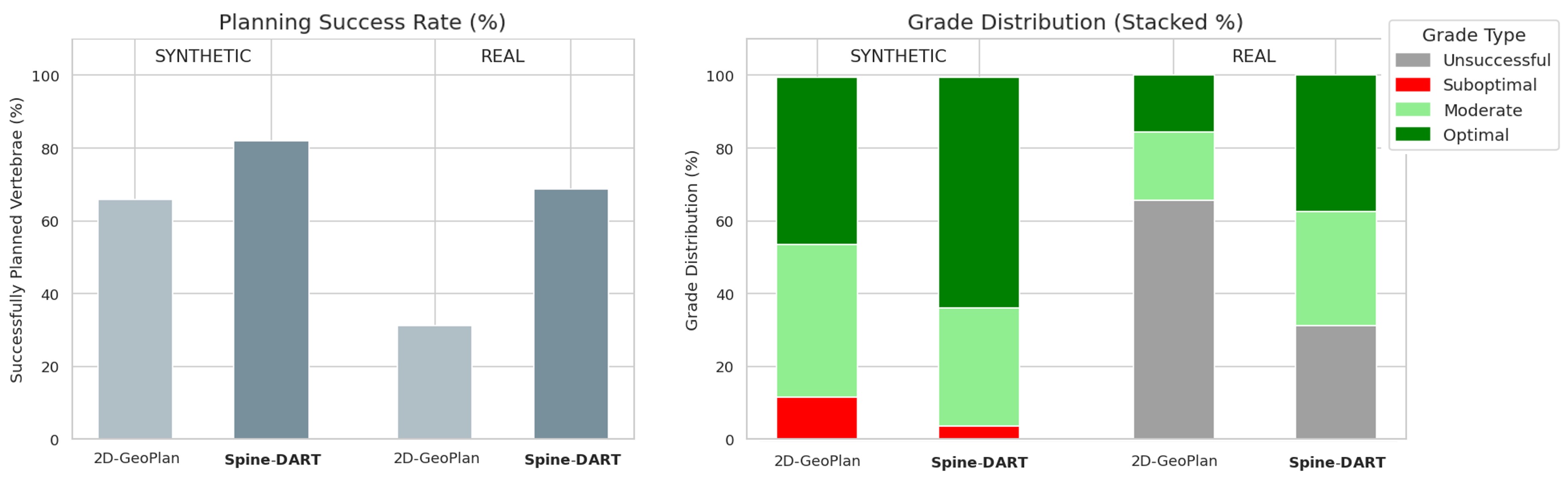}
    \caption{Left: Percentage of successfully planned vertebrae for the 2D-GeoPlan and \textbf{Spine-DART} models, shown separately for the synthetic dataset and the real X-ray dataset. Right: Stacked grade distribution of all evaluated cannula paths (200 synthetic paths and 32 real X-ray paths), illustrating the proportion of unsuccessful, suboptimal, moderate, and optimal plans assigned by a clinical expert. Together, the plots summarize overall planning success and qualitative path assessment for both models across synthetic and real data.}
    \label{fig:success_rate_results}
\end{figure}

\begin{figure}
    \centering
    \includegraphics[width=1\textwidth]{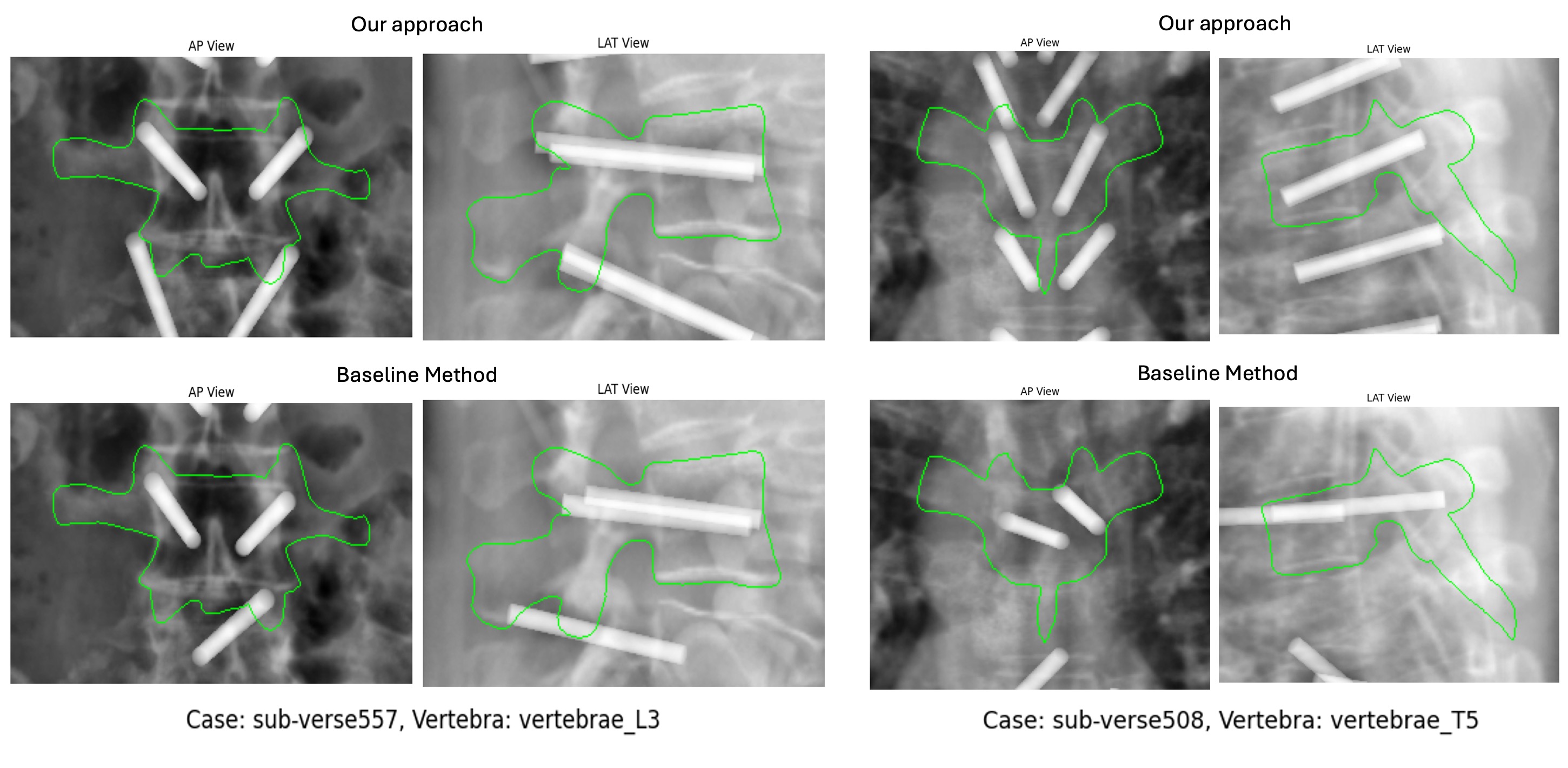}
    \caption{Anteroposterior (AP) and lateral (LAT) DRRs for two vertebral samples from VerSe19. Cannula placements are shown for 2D-GeoPlan (top) and Spine-DART (bottom).}
    \label{fig:eval_paths}
\end{figure}

Since there is no single correct solution for the transpedicular trajectory, we did not directly compare the reconstructed paths to manual annotations. Instead, we randomly selected 100 vertebrae that had successful reconstructions from both the baseline and our proposed method. These reconstructed paths, along with those from a cadaver specimen, were evaluated by a medical resident in Interventional Radiology at Johns Hopkins School of Medicine. For the assessment, we simulated orthogonal DRRs with the reconstructed cannulas rendered as metallic objects within the scene (Figure~\ref{fig:eval_paths}). The clinician graded the placement accuracy of each cannula on a 3-point scale: 0 — Suboptimal, 1 — Moderate/requires adjustment, and 2 — Optimal placement. Figure~\ref{fig:success_rate_results} summarizes the grade distribution across the 100 vertebrae (200 cannulas in total) based on the expert's evaluation.

\subsection{Discussion and Conclusion}

Accurate 3D surgical planning from intraoperative 2D X-rays remains a key challenge in enabling robotic assistance for procedures where preoperative CT is not available. Our proposed optimization-based VA planning framework —Spine-DART— addresses this challenge by generating clinically viable transpedicular trajectories from just two intraoperative X-rays, even under diverse and non-standard imaging conditions. 

The observed reconstruction accuracy —comparable to state-of-the-art deep learning models trained on fixed-view protocols— highlights the strength of our differentiable optimization framework in handling arbitrary intraoperative imaging. While DL such as ReVerteR offer slightly better reconstruction metrics under idealized conditions, their reliance on fixed geometries and lack of built-in anatomical correspondences limit downstream utility for path planning. In contrast, Spine-DART trades off minor decreases in reconstruction precision for greater flexibility, interpretability, and planning-readiness.

By leveraging SSMs and differentiable rendering, Spine-DART introduces anatomical consistency and interpretability that purely data-driven approaches often lack. This enables direct propagation of semantic labels —such as transpedicular trajectories— from the reconstructed surface. During qualitative evaluation by a clinical expert, our planned cannula placements, derived from arbitrary X-ray views, were more frequently rated as “clinically viable” than those produced by a 2D-3D triangulation baseline. These results suggest that anatomical priors and gradient-based refinement not only produce plausible reconstructions but also support safer, more informed planning.

Importantly, this framework extends beyond line-based path planning. The availability of dense anatomical correspondences enables the propagation of complex, multi-segment trajectories, which are essential for more advanced procedures like curved Balloon Kyphoplasty. Moreover, the differentiable formulation naturally supports future extensions, including biomechanical constraints and multi-objective optimization to balance surgical access, safety margins, and implant positioning. In addition, while Spine-DART is demonstrated using only two X-rays, the framework can be readily extended to incorporate additional views, further improving reconstruction accuracy and planning robustness when clinically available.

Nevertheless, the pipeline has limitations. Its modular nature means that overall performance depends on the accuracy of several upstream components, including landmark detection, SSM fitting capability, and image similarity estimation. In particular, failures to detect sufficient overlapping landmarks across views can compromise initialization. While backup strategies and semantic priors help mitigate this, incorporating probabilistic keypoint confidence or attention mechanisms could improve robustness. Additionally, the very strength of the SSM —its ability to model anatomically plausible shapes— can limit its accuracy when fitting to low-prevalence morphologies, such as compressed or fractured vertebrae, which are precisely the cases targeted by vertebral augmentation. Future work will focus on improving shape modeling for such cases, including per-level vertebral representations rather than region-specific templates. Lastly, although current optimization runtimes are suitable for planning, integrating surrogate models could enable real-time reconstruction for interactive surgical use.

Ultimately, this work demonstrates that accurate 3D path planning for robot-assisted vertebroplasty is feasible from arbitrary bi-plane X-rays using a differentiable rendering framework. By eliminating the need for preoperative CT and embracing real-world imaging variability, Spine-DART offers a practical and clinically viable solution for intraoperative guidance. However, while our results indicate a clear advancement over conventional baselines, some trajectories still fail, and many would benefit from manual adjustment, highlighting the need for further refinement before the system can be considered fully automatic. Future work will focus on improving anatomical modeling —particularly for challenging cases such as fractured vertebrae— developing failure detection and recovery mechanisms to flag suboptimal planning outcomes, and enabling real-time deployment in the OR. Clinical translation efforts will aim to integrate the system seamlessly into routine workflows, while also extending the framework to support more complex interventions such as curved trajectory planning in Balloon Kyphoplasty.

\section*{Funding}
This work was funded in part by NIH R01 EB036341 and Johns Hopkins Internal Funds.

\section*{Conflict of Interest Statement}
The authors declare that the research was conducted in the absence of any commercial or financial relationships that could be construed as a potential conflict of interest.



\bibliographystyle{Frontiers-Harvard} 
\bibliography{test}

\end{document}